\documentclass{article}

\usepackage{microtype}
\usepackage{graphicx}
\usepackage{subfigure}
\usepackage{booktabs} % for professional tables

\usepackage{hyperref}
\usepackage[font={tiny}]{caption}
\usepackage{array}
\newcolumntype{?}{!{\vrule width 1pt}}

\usepackage[accepted]{icml2023-diffxyz}

\usepackage{amsmath}
\usepackage{amssymb}
\usepackage{mathtools}
\usepackage{amsthm}

\usepackage[capitalize,noabbrev]{cleveref}

\theoremstyle{plain}

\theoremstyle{definition}

\theoremstyle{remark}

\DeclareMathOperator*{\argmax}{arg\,max}
\DeclareMathOperator*{\argmin}{arg\,min}
\usepackage[textsize=tiny]{todonotes}

\icmltitlerunning{Learning Observation Models with Incremental Non-Differentiable Graph Optimizers in the Loop for Robotics State Estimation}

\begin{document}

\twocolumn[
\icmltitle{Learning Observation Models with Incremental Non-Differentiable Graph Optimizers in the Loop for Robotics State Estimation}



\icmlsetsymbol{equal}{*}

\begin{icmlauthorlist}
\icmlauthor{Mohamad Qadri}{yyy}
\icmlauthor{Michael Kaess}{yyy}

\end{icmlauthorlist}

\icmlaffiliation{yyy}{Robotics Institute, Department of Computer Science, Carnegie Mellon University, Pittsburgh, United States}

\icmlcorrespondingauthor{Mohamad Qadri}{mqadri@andrew.cmu.edu}

\icmlkeywords{Machine Learning, ICML}

\vskip 0.3in
]



\printAffiliationsAndNotice{\icmlEqualContribution} % otherwise use the standard text.
\begin{figure*}[h!]
    \centering
    \includegraphics[width=1\textwidth]{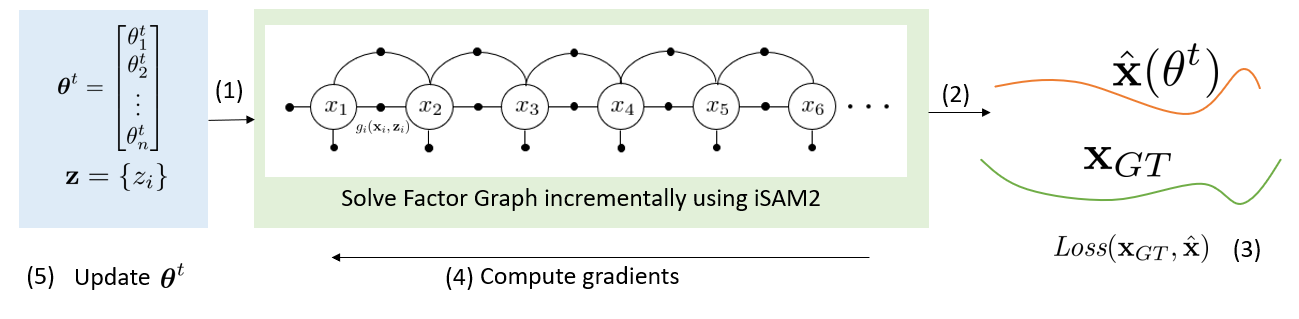}
    \caption{The parameter vector $\boldsymbol{\theta}^t$ serve as input to the least squares problem which is solved by iSAM2. Gradients of the solution with respect to the parameters are computed and used to update $\boldsymbol{\theta}^{t+1}$}
    \label{fig:VDM}
\end{figure*}

\begin{abstract}
We consider the problem of learning observation models for robot state estimation with incremental non-differentiable optimizers in the loop. Convergence to the correct belief over the robot state is heavily dependent on a proper tuning of observation models which serve as input to the optimizer. We propose a gradient-based learning method which converges much quicker to model estimates that lead to solutions of much better quality compared to an existing state-of-the-art method as measured by the
tracking accuracy over unseen robot test trajectories.
\end{abstract}
\vspace{-8mm}
\section{Introduction}
\vspace{-1mm}
Robot state estimation is the problem of inferring the state of a robot (a set of geometric or physical quantities such as position, orientation, contact forces etc.) given sensor measurements. The problem is typically formulated as Maximum a Posteriori (MAP) inference over factor graphs where each node (robot state) is connected to other states via soft constraints or potentials (factors) which are distilled from sensor measurements. Given a factor graph, a potential way to perform the inference step is to convert it to a chordal Bayes Net using an exact inference technique (i.e. variable elimination) and then in turn, convert the Bayes Net to a tree like structure (i.e. junction tree) where inference can be made easier. However, this procedure can become ill-suited for real-time state estimation especially if the number of variables in the problem increases with time (i.e. as the robot navigates the environment). The \textit{Bayes tree} \cite{kaess2011bayes} was introduced to tackle this problem. It is a tree structure
 where nodes are formed from the maximal cliques of a chordal Bayes net. The Bayes tree is directed and maintains the conditional independences described by the original Bayes net. In addition, it allows for fast incremental inference which corresponds to simple tree editing.

iSAM2 \cite{kaess2012isam2} is an optimizer that leverages the Bayes tree to solve incremental (i.e. as the number of factors $N$ grows continuously) inference problems formulated as a factor graph. In essence, it solves problems of the form:
\begin{align}
    \mathbf{x}^* = \argmin_\mathbf{x} \sum_{i=1}^N \frac{1}{2} ||g_i(\mathbf{x}_i) - z_i ||_{\theta_i}^2
    \label{maineq}
\end{align} 
or equivalently the MAP inference problem:
\begin{align}
    \mathbf{x}^* = \argmax_\mathbf{x} \prod_i^N \phi_i(\mathbf{x}_i)
    \label{mainmap}
\end{align} 
where $\phi_i$ are the potentials assumed to take the form $\phi_i\!\propto\! \exp\left( -\frac{1}{2} ||g_i(\mathbf{x}_i) -z_i)||_{\theta_i}^2 \right)$, $\mathbf{x}$ is the vector of unknown states, $\mathbf{x}_i$ a subset of $\mathbf{x}$, and $\mathbf{z} = \{z_i\}$ are the sensor measurements. Batch solvers such as Levenberg-Marquardt perform the typical linearize-solve loop which quickly becomes prohibitively expensive as the size of the problem grows. iSAM2, on the other hand, leverages the Bayes tree and uses \textit{fluid relinarization} and \textit{partial state updates} making it the de-facto optimizer for online smoothing-based state estimation problems in robotics. However, these features also prevent iSAM2 from being differentiable (i.e. dependence on relinearization thresholds, inherent non-differentiable operations such as removal and re-insertion of tree nodes.) 

On the other hand, the quality of the solution of eq.\ref{maineq} is largely dependent on a proper selection of the observation models $\{\theta_i\}$ which parameterize the joint or conditional distributions over states and measurements. However, due the non-differentiability of iSAM2, current methods to tune these parameters are generally sampling-based which are slow and may converge to poor optimas.
% Hence, current methods to tune the observation models  $\theta_i$ in eq. \ref{maineq} (with iSAM2 in the loop),  which heavily affect the quality of the solution, is either manual or using 
In this work, we propose a gradient-based optimization-based approach to tune $\{\theta_i\}$ with iSAM2 in the loop using a direct tracking error loss. We compare our method with a state-of-the-art sampling-based method on a synthetic robot navigation example and show that our procedure converges order-of-magnitude faster to a better solution as measured by the tracking accuracy over unseen robot test trajectories.

% from data by taking gradient of the solution wrt to parameters with iSAM2 in the loop irrespective to the non-differentiable components of the optimizer. 

\vspace{-2mm}
\section{Related Work}
Early state estimation techniques such as the family of Kalman Filters \cite{julier2004unscented, thrun2001robust} rely on the Markov assumption to enable real-time performance. Recent algorithms have been proposed to make these filters differentiable \cite{kloss2021train, sun2016learning, haarnoja2016backprop}. However, the inherent reliance on the Markov assumption and the inability to re-linearize past states can lead to convergence to poor solutions. Hence, state-of-the-art robotic state estimation algorithms moved to factor graph-based solutions which encode the inherent temporal structure avoiding the need to marginalize past states and providing methods to relinearize past estimates \cite{kaess2012isam2, 9982178}.  However, observation models on factor graphs are either pre-specified and fixed \cite{lambert2019joint, engel2014lsd} or learned using surrogate losses independent of the graph optimizer or final tracking performance \cite{sodhi2021learning, sundaralingam2019robust, czarnowski2020deepfactors}. Methods that differentiate through the $\argmin$ operator in eq. \ref{maineq} by unrolling the optimizer can be used to learn these models \cite{yi2021differentiable, gradslam, bechtle2021meta}. However, these techniques are typically sensitive to hyperparameters such as the number of unrolling steps \cite{amos2020differentiable} in addition to suffering from vanishing as well as high bias and variance gradients. Recently, a novel method \textit{LEO} \cite{sodhi2022leo} took advantage of the probabilistic view offered by iSAM2 (as a solver of eq.\ref{mainmap}) to provide a way to learn observation models, with iSAM2 in the loop, by minimizing a novel tracking error. In essence, at every training iteration LEO samples trajectories from the posterior distribution (a joint Gaussian distribution over the states) and the deviation with respect to the ground truth trajectory is minimized using an energy-based loss. In this work, we use LEO as our baseline.
\vspace{-2mm}
\section{Method}
In this work, we view the incremental optimizer (iSAM2) as a function $f: \mathcal{X} \times \boldsymbol{\theta} \rightarrow \mathcal{X}$ which takes initial estimates of the state $\mathbf{x} \in \mathcal{X}: \mathcal{M}_1 \times \hdots \times \mathcal{M}_n$ at timestep $t$ (the number of states increases with time), as well as parameters $\boldsymbol{\theta} = \{\theta_i \; | \; \theta_i \in \mathcal{S}^{n_{i}}_{++}\}$ and returns an estimate of the optimal state $\mathbf{x}_* \in \mathcal{X}$ after performing $N$ update steps. Here, $\mathcal{M}_i$ is a Lie Group (for example the special Euclidean group $SE(n)$) and $\mathcal{S}^{n_{i}}_{++}$ is the set of $n_i \times n_i$ positive definite matrices . 

Our goal is to learn the parameters $\{\theta_i\} $ using gradient-based methods from observed ground truth robots trajectories $\mathbf{x}_{GT}$. Specifically, We consider the following inner-outer optimization procedure:
\begin{align}
    % & \text{Inner Loop: } \; f(\boldsymbol{\theta}; \mathbf{x}^0) =  \argmin_\mathbf{x} G(\mathbf{x},\boldsymbol{\theta}; \mathbf{z}, \mathbf{x}^0) = \hat{\mathbf{x}}(\boldsymbol{\theta})  \nonumber \\
    & \text{Inner Loop: } \; \hat{\mathbf{x}}(\boldsymbol{\theta)} =  \argmin_\mathbf{x} H(\mathbf{x},\boldsymbol{\theta}; \mathbf{z})  \nonumber \\
    & \quad \quad \quad \quad \quad  \quad \; \; \; \; \, \, = \argmin_\mathbf{x} \sum_i \frac{1}{2}||g_i(\mathbf{x}) - z_i||^2_{\theta_i} 
    \label{innerLoop} \\
    & \text{Outer Loop:} \; \min_{\boldsymbol{\theta}} \mathcal{L}(\hat{\mathbf{x}}(\boldsymbol{\theta)}, \mathbf{x}_{GT})
    \label{outerloop}
\end{align}
Note that the gradient $\frac{\partial L}{\partial \boldsymbol{\theta}}$, requires an estimate of  $\frac{\partial \hat{\mathbf{x}}}{\partial \boldsymbol{\theta}}$. Although iSAM2 includes different non-differentiable operations, the gradient of the solution with respect to the parameter vector $\frac{\partial \hat{\mathbf{x}}}{\partial \boldsymbol{\theta}}$ exists and can be computed by the implicit function theorem \cite{dontchev2009implicit} as done in existing work in convex optimization \cite{amos2017optnet, agrawal2019differentiable}. Note that the original theorem consider functions operating on vector spaces. However, the theorem can readily be extended to other manifolds by applying the appropriate group operations. Let $h(\mathbf{x}, \boldsymbol{\theta}) := \frac{\partial H}{\partial \mathbf{x}}$.

\textbf{\textit{The Implicit Function Theorem:}}
\newline 
\textit{
Let $\mathbf{\hat{x}}(\boldsymbol{\theta}) := \{\mathbf{x} \; | \; h(\mathbf{x}, \boldsymbol{\theta})=0\}$ where $\mathbf{x} \in \mathcal{X}$ and $\boldsymbol{\theta} = \{ \theta_i  \; | \; \theta_i \in \mathcal{S}^n_{++} \}$. Let $h$ be continuously differentiable in the neighborhood of $(\mathbf{\hat{x}}, \boldsymbol{\theta})$ namely $\nabla_{\mathbf{x}} h(\mathbf{\hat{x}}(\boldsymbol{\theta}), \boldsymbol{\theta})$ be nonsingular. Then:
}
\begin{align}
    \frac{\partial f}{\partial \boldsymbol{\theta}}=\frac{\partial \mathbf{\hat{x}}(\boldsymbol{\theta})}{\partial \boldsymbol{\theta}} = - \left(\frac{\partial h(\mathbf{\hat{x}}(\boldsymbol{\theta}), \boldsymbol{\theta})}{\partial \mathbf{x}} \right)^{-1} \frac{\partial h(\mathbf{\hat{x}}(\boldsymbol{\theta}), \boldsymbol{\theta})}{\partial \mathbf{\boldsymbol{\theta}}}
    \label{gradientIFT}
\end{align}
The gradient in eq. \ref{gradientIFT} can be derived and computed analytically. However, we note that since the size of parameter vector $\boldsymbol{\theta}$ is typically small (for example, each $\theta_i$ has a maximum of $6$ free parameters when working with elements in $SE(2)$), numerical differentiation proved to be efficient especially when coupled with parallelization on CPU.
\begin{figure*}[h!]
    \centering
    \includegraphics[width=1\textwidth, height=2.5in]{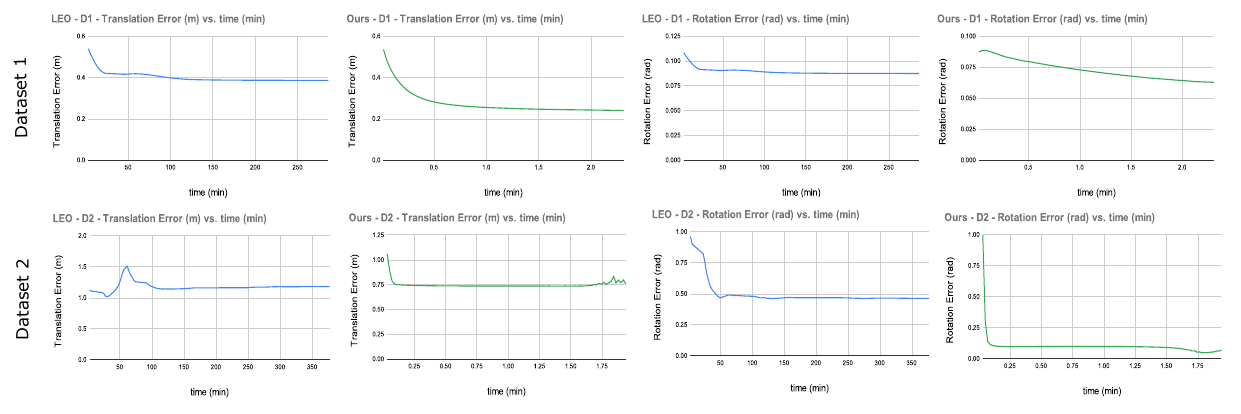}
    \caption{Convergence as a function of training time measured in minutes. Both methods were run for 100 training iterations. Our method converges after a few seconds while each iteration of LEO takes approximately 2.5 minutes. Additionally, our method converges to parameters that track the training trajectories more accurately.}
    \label{fig:Convergencespeed}
\end{figure*}

\begin{figure*}[h!]
    \centering
    \includegraphics[width=1\textwidth, height=1.25in]{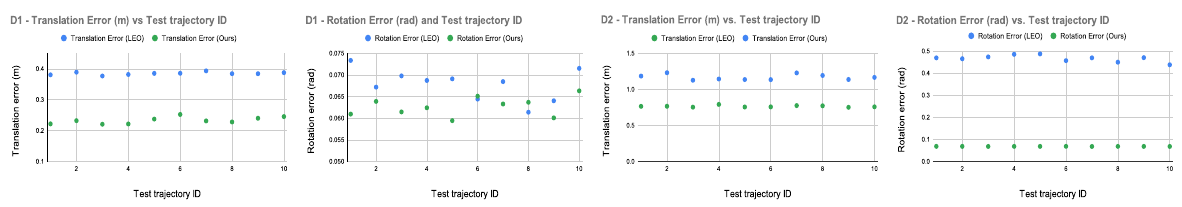}
    \caption{Accuracy on the test set. Our method converges to parameters that generalize and track the test trajectories more accurately. }
    \label{fig:testError}
\end{figure*}

\subsection{Numerical Jacobians over Lie Groups}
\subsubsection{Jacobians on Vector Spaces}
Given a multivariate function $f: \mathbb
{R}^m \rightarrow \mathbb{R}^n$, the Jacobian is defined as the $n \times m$ matrix:
\begin{align}
    \frac{\partial f(\mathbf{x})}{\partial \mathbf{x}} = \begin{bmatrix}
        \frac{\partial f_1}{\partial x_1} & \hdots & \frac{\partial f_1}{\partial x_m} \\
        \vdots & \vdots & \vdots \\ 
        \frac{\partial f_n}{\partial x_1} & \hdots & \frac{\partial f_n}{\partial x_m} \\        
    \end{bmatrix}
\end{align}
where $\frac{\partial f}{\partial x_i} = \lim\limits_{h \rightarrow 0}\frac{f(\mathbf{x} + h \mathbf{e}_i) - f(\mathbf{x})}{h}$ and 
 $\mathbf{e}_i$ is the $i$th standard basis of $\mathbb{R}^m$.
\subsubsection{Left Jacobians on Lie Groups}
We can similarly define the left Jacobian of functions acting on manifolds $f: \mathcal{N} \rightarrow \mathcal{M}$ as the linear map from the Lie algebra $T_\epsilon(\mathcal{N})$ of $\mathcal{N}$ to $T_\epsilon(\mathcal{M})$, the Lie algebra of $\mathcal{M}$:
\begin{align}
    & \frac{{}^{\epsilon}Df(\mathcal{X})}{D\mathcal{Y}} = \lim\limits_{\tau \rightarrow 0 } \frac{f(\tau \oplus \mathcal{Y}) \ominus f(\mathcal{Y})}{\tau} \\ 
    & \quad \quad \quad \; \; \;  = \lim\limits_{\tau \rightarrow 0 } \frac{\text{Log}(f(\text{Exp}(\tau) \circ \mathcal{Y}) \circ f(\mathcal{Y})^{-1})}{\tau}
\end{align}
where $\mathcal{Y} \in \mathcal{N}$, $\tau$ is a small increment defined on $T_\epsilon(\mathcal{N})$. The Log operator map elements from a Lie Group to its algebra while the Exp operator map elements from the algebra to the group.  $\oplus$,  $\ominus$, and $\circ$ are the plus, minus, and composition operators respectively \cite{sola2018micro} where:
\begin{align}
    & \tau \oplus \mathcal{Y} = \text{Exp}(\tau) \circ \mathcal{Y} \\ 
    & \tau = \mathcal{Y}_1 \ominus   \mathcal{Y}_2 = \text{Log}(\mathcal{Y}_1 \circ \mathcal{Y}_2^{-1}); \; \mathcal{Y}_1,\mathcal{Y}_2 \in \mathcal{N}
\end{align}
In this work, $\mathcal{N} = \mathcal{S}^{n_1}_{++} \times \hdots \times \mathcal{S}^{n_m}_{++}$ and $\mathcal{M} = \mathcal{X}$. Additionally, we assume that each vector $\theta_i$ are the square root elements of some corresponding diagonal positive definite matrix $\Sigma_i \in  \mathcal{S}^{n_i}_{++}$. i.e., we define the following map:
\begin{align}
    \theta_i = \left(\text{diag}^{-1}(\Sigma_i)\right)^2 \in \mathbb{R}^{n_i}
\end{align}
Hence, $\tau \in \mathbb{R}^{n_i}$ and the operator $\oplus$ is the standard addition on vector space $\mathbb{R}^{n_i}$. 
\vspace{-3mm}
\subsection{Tracking Loss}
Let a parameter estimate $\boldsymbol{\theta}^t \in \mathbb{R}^m$. The outer loss is the regularized mean squared error between the estimated trajectory $\mathbf{\hat{x}} (\boldsymbol{\theta}^t)$ and the ground truth $\mathbf{x}_{GT}$. Let $\mathcal{D}$ be the training set, $T$ be the total number of states in $\mathbf{x}_{\text{GT}}$, $D$ be the sum of the lie algebra dimensions of all states, and $\lambda \in \mathbb{R}$:
\begin{align}
&\!\!\! \mathcal{L}(\boldsymbol{\theta})=
% \begin{bmatrix}
%    \textbf{vec}(\frac{1}{2} || \mathbf{\hat{x}} (\boldsymbol{\theta}^t_{11})\ominus \mathbf{x}_{GT}||_2^2) \\
%    \vdots \\
%    \textbf{vec}(\frac{1}{2} || \mathbf{\hat{x}} (\boldsymbol{\theta}^t_{NM})\ominus \mathbf{x}_{GT}||_2^2) 
% \end{bmatrix} + \lambda ||\boldsymbol{\theta}^t||_2^2
\frac{1}{2|\mathcal{D}|} \sum_{j=1}^\mathcal{|D|}|| \textbf{vec}(\mathbf{\hat{x}}^j (\boldsymbol{\theta}^t)\ominus \mathbf{x}_{GT})||_2^2+ \lambda ||\boldsymbol{\theta}^t||_2^2 \\
&\!\!\! \!\frac{\partial \mathcal{L}}{\partial \boldsymbol{\theta}}\!=\!\frac{1}{|\mathcal{D}|} \sum_{j=1}^{|\mathcal{D}|} S(\mathbf{\hat{x}}^j(\boldsymbol{\theta}^t))^T
\cdot \underbrace{\textbf{vec}(\mathbf{\hat{x}}^j (\boldsymbol{\theta}^t)\ominus \mathbf{x}_{GT})}_{\in \mathbb{R}^{TD}} + 2\lambda \boldsymbol{\theta}^t
\end{align}
where $\mathbf{vec}$ is the vectorization operator and $S(\boldsymbol{\hat{x}}(\boldsymbol{\theta}^t)) \in \mathbb{R}^{TD \times m}$ is a matrix such that each row $r$ is equal to:  
\begin{align}
    &S(\mathbf{\hat{x}}^j(\boldsymbol{\theta}^t))_{r} = \textbf{vec}\left(\frac{\partial \mathbf{\hat{x}}^j}{\partial \theta_{ij}}\right) \nonumber\\
    & \quad \quad \quad \quad \; \;= \lim\limits_{\tau_{ij} \rightarrow 0} \frac{\textbf{vec}(\text{Log}(\mathbf{\hat{x}}^j(\Tilde{\boldsymbol{\theta}}^t) \circ \mathbf{\hat{x}}^j(\boldsymbol{\theta}^t)^{-1}))}{\tau_{ij}}
    \label{fd}
\end{align}
where $\Tilde{\theta}^t_{ij} = \theta^t_{ij} + \tau_{ij}$
 and $\Tilde{\boldsymbol{\theta}}^t = \boldsymbol{\theta}^t$ otherwise. By the implicit function theorem, the gradient $\frac{\partial \mathbf{\hat{x}}}{\partial \theta_{ij}}$ exists and is estimated in eq. \ref{fd} using finite differencing by perturbing the parameter $\theta_{ij}$ by $\mathbf{\tau}_{ij}$ . Parameters are then updated using gradient descent with learning rate $\alpha$:
\begin{align}
\boldsymbol{\theta}^{t+1} = \boldsymbol{\theta}^t - \alpha \cdot \frac{\partial \mathcal{L}}{\partial \boldsymbol{\theta}}
\label{GD}
\end{align}
\vspace{-10mm}
\section{Results}
\begin{table}[h!]
    \renewcommand\thetable{I}
    \vspace{-2mm}
    \caption{Comparison between LEO and Ours (final training runtime and error, test error statistics).}%\textit{BF} indicates brute force matching.}
    \begin{center}
    \resizebox{1\linewidth}{!}{
        \begin{tabular}{c|c|c|c|c}
            \toprule
            \textbf{} & LEO ($D_1$) & Ours ($D_1$) & LEO ($D_2$) & Ours ($D_2$)  \\
            \hline
            Num training iterations & 100 & 100 & 100 & 100 \\
            Parallelization enabled during training & yes & \textbf{no} & yes & \textbf{no} \\
            Time per training iteration (min) & 2.863 & $ \mathbf{0.025}$ & $3.773$ & $\mathbf{0.022}$ \\
            \hline
            Num of training trajectories & 5 & 5 & 5 & 5 \\
            Length of training trajectories & 300 & \textbf{100} & 300 & \textbf{100} \\
            Avg training RMS translation error (m) & 0.385 & \textbf{0.236} & 1.179 & \textbf{0.730} \\
            Avg training RMS rotation Avg error (rad) & 0.087 & \textbf{0.062} & 0.467 & \textbf{0.051} \\
            \hline
            Num test trajectories & 20 & 20 & 20 & 20   \\
            Length of test trajectories & 300 & 300  & 300 & 300 \\
            Avg test RMS translation error (m) & 0.384 & \textbf{0.238} & 1.153 & \textbf{0.733}  \\
            Avg test RMS rotation error (rad) & 0.067 & \textbf{0.063} & 0.463 & \textbf{0.050}  \\
            \toprule
        \end{tabular}
    }
    \label{table:summaryresult}
    \end{center}
    \vspace{-3mm}
\end{table}

We use two synthetic planar (i.e. in $SE(2)$) robotic navigation datasets $\textit{D}_1$ and $\textit{D}_2$ consisting of GPS and odometric measurements. Each dataset uses a different set of parameters $ \{ \theta_{\text{GPS}}, \theta_{\text{odom}} 
 \}_{\textit{D}_i} $  to generate $N$ trajectories (more details in appendix \ref{appendixA}). 
 We use 5 training ground truth trajectories of length\footnote{Length refers to the number of nodes in the graph optimizer} 300 to train  LEO\footnote{We use the official implementation of LEO by Paloma et al.} and 5 training trajectories of length 100 to train our method. We use a set of 20 unseen trajectories (all of length 300) to test the tracking performance given our final learned parameters (the testing set is fixed for both methods).  For LEO, we enable multithreaded parallel trajectory sampling while do not use any form of parallelization with our technique. 
 % and an NVIDIA RTX 3090 GPU, an Intel Core i9-10900K, and 32GB of RAM to train it.
 
 At each training iteration, iSAM2 optimizes the inner loop (eq. \ref{innerLoop}) objective for both methods:
 \begin{align}
     & \mathbf{\hat{x}}\!=\! \argmin_\mathbf{x} H( \boldsymbol{\theta}, \mathbf{x}; \mathbf{z})\!=\! \argmin_\mathbf{x}\! 
     \frac{1}{2}\! \sum_i\biggl(\!||g^{\text{gps}}(x_i) \!-\! z_{i}^{\text{gps}}||_{\theta_\text{gps}}^2   \nonumber \\
     &\quad \quad \quad \quad \quad +  \frac{1}{2}||g^{\text{odom}}(x_{i-1}, x_i) - z_{i-1, i}^{\text{odom}}||_{\theta_\text{odom}}^2 \biggr)
 \end{align}
Our method then updates the parameters $\boldsymbol{\theta}$ using eq.\ref{GD} while LEO minimizes the following energy-based loss:
\begin{align}
    &\!\!\!\!\!\!\!\!\mathcal{L}(\theta)\!=\!\frac{1}{|\mathcal{D}|} \! \sum_{(\mathbf{x}_{\text{gt}}^j, \mathbf{z}^j) \in \mathcal{D}}\!\!\!\!\!
E(\boldsymbol{\theta},\mathbf{x}_{\text{gt}}^j;\mathbf{z}^j)\!+\!\log\! \int_\mathbf{x} e^{-E(\boldsymbol{\theta}, \mathbf{x};\mathbf{z}^j)} d\mathbf{x}
\label{leoloss}
\end{align}
where $(\mathbf{x}_{\text{gt}}^j, \mathbf{z}^j)$ is a ground truth sample  from the training set $\mathcal{D}$, the energy $E( \boldsymbol{\theta}, \mathbf{x}; \mathbf{z}) := H(\boldsymbol{\theta}, \mathbf{x}; \mathbf{z})$, and the integral is over the space of trajectories. Fig. \ref{fig:Convergencespeed} shows the training time and root mean squared error (RMSE) at each iteration. Fig. \ref{fig:testError} shows the RMSE on the test set, and table \ref{table:summaryresult} gives a summary of the results. We note that our method converges to parameters that lead to better tracking accuracy on all unseen test trajectories. In addition, while LEO needs to generate samples from a high dimensional probability distribution during training which is a time consuming process, our method generates gradients by directly comparing deviation from the training trajectories leading to order-of-magnitude faster training time. We additionally analyze the performance of both methods as a function of the training set size and provide the results in appendix \ref{additionalresults}.
\vspace{-2mm}
\section{Conclusion and Future Work}
We presented a gradient-based learning algorithm which estimates observation models with non-differentiable optimizers (iSAM2) in the loop for robotic state estimation. While current state of the art algorithms require sampling trajectories from the posterior distribution to bypass the non-differentiability of the optimizer, our technique learns parameters by formulating the problem as a bilevel optimization procedure where gradients are generated through numerical differentiation. 

For future work, we want to extend our algorithm to learn parameters $\{\theta_i\}$, with iSAM2 in the loop, that are themselves functions of observations i.e. $\theta_i(z_i, \phi_i)$ where $\phi_i$ can, for example, be the weights of a jointly trained neural network. Indeed, the outputs of the network can be perturbed to approximate $\frac{\partial\mathbf{\hat{x}}}{\partial \theta_i}$ as proposed in this work and then simply chained with the gradient $\frac{\partial \theta_i}{\partial \phi}$ (obtainable from existing auto-differentiation packages such as PyTorch) to get the gradient of the optimized output trajectory with respect to network weights. We plan to compare the sample efficiency, training time, and generalization performance of our method against different baselines. Finally, we plan to train our algorithm on real robotics data and deploy our state estimator on real robotics platforms.

\bibliography{main}
\bibliographystyle{icml2023}

%%%%%%%%%%%%%%%%%%%%%%%%%%%%%%%%%%%%%%%%%%%%%%%%%%%%%%%%%%%%%%%%%%%%%%%%%%%%%%%
%%%%%%%%%%%%%%%%%%%%%%%%%%%%%%%%%%%%%%%%%%%%%%%%%%%%%%%%%%%%%%%%%%%%%%%%%%%%%%%
% APPENDIX
%%%%%%%%%%%%%%%%%%%%%%%%%%%%%%%%%%%%%%%%%%%%%%%%%%%%%%%%%%%%%%%%%%%%%%%%%%%%%%%
%%%%%%%%%%%%%%%%%%%%%%%%%%%%%%%%%%%%%%%%%%%%%%%%%%%%%%%%%%%%%%%%%%%%%%%%%%%%%%%

\newpage
\appendix
\onecolumn
\section{Appendix}
\subsection{Additional Details on Experimental Setup}
\label{appendixA}

\begin{figure*}[h!]
    \centering
    \includegraphics[width=1\textwidth]{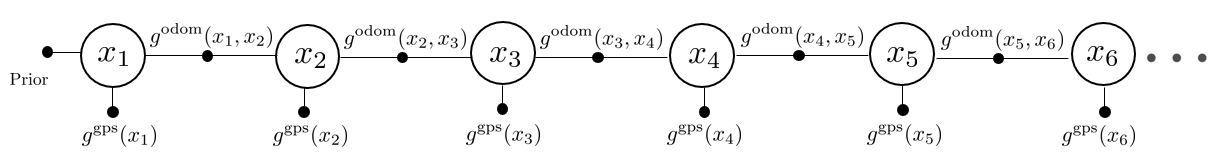}
    \caption{The factor graph used to generate the synthetic robot navigation trajectories.}
    \label{fig:appendixpic}
\end{figure*}
Figure \ref{fig:appendixpic} shows the structure of the factor graph used to generate the synthetic robot navigation trajectories. A unary GPS factor $g^{\text{gps}}(x_i)$ is added to each pose $x_i$ while a binary odometry factor $g^{\text{gps}}(x_i, x_{i-1})$ is specified between poses. To simulate realistic robot navigation trajectories for each of datasets $D_1$ and $D_2$, Gaussian noise $\sim \mathcal{N}(0, \theta_{\text{odom}})$ is injected to ground truth relative odometry measurements while Gaussian noise $\sim \mathcal{N}(0, \theta_{\text{gps}})$ is added to absolute ground truth GPS measurements.
\vspace{-2mm}
\subsection{Additional Experiments}
\label{additionalresults}
We  analyze our method and LEO as a function of the training set size. 
Tables \ref{table:detailD1} and \ref{table:detailD2} show the average training and testing RMS errors on datasets $D_1$ and $D_2$ respectively. We note that our method possesses a much faster training time as well as learns parameters that generalize better to unseen test trajectories regardless of the training set size. 
% On the other hand, we also note that accuracy on the testing set does not improve with additional training samples due to the small parameter space.
\begin{table}[h!]
    \vspace{-4mm}
    \renewcommand\thetable{II}
    \caption{Dataset $D_1$ - Comparison between LEO and Ours (final training runtime and error, test error statistics).}%\textit{BF} indicates brute force matching.}
    \begin{center}
    \resizebox{1\linewidth}{!}{
        \begin{tabular}{c|c|c||c|c||c|c||c|c||c|c}
            \toprule
            \textbf{Num trajectory in training set} & \multicolumn{2}{c||}{1} & \multicolumn{2}{c||}{5} & \multicolumn{2}{c||}{10} & \multicolumn{2}{c||}{20} & \multicolumn{2}{c}{30} \\
            \hline
            \textbf{Algorithm} & LEO & Ours & LEO & Ours & LEO & Ours & LEO & Ours & LEO & Ours \\
            \hline
            Num training iterations & 100 & 100 & 100 & 100 & 100 & 100 & 100 & 100 & 100 & 100  \\
            Parallelization enabled during training & yes & \textbf{no} & yes & \textbf{no} & $\text{yes}^*$ & \textbf{no} & $\text{yes}^*$ & \textbf{no} & $\text{yes}^*$ & no\\
            Time per training iteration (min) & 0.937 & $\mathbf{0.005}$ & $2.863$  & \textbf{0.025} & 4.370 & \textbf{0.045} & 4.899 & \textbf{0.091} & 7.56 & \textbf{0.136} \\
            \hline
            Average training RMS translation error (m) & 0.379  & \textbf{0.257} & 0.385 &  \textbf{0.236} &  0.383 &  \textbf{0.243} &  0.386 & \textbf{0.243} & 0.385  & \textbf{0.236}\\
            Average training RMS rotation Avg error (rad) &   0.071 & \textbf{0.060} &  0.087 & \textbf{0.062} & 0.076 & \textbf{0.063} & 0.079 &  \textbf{0.068} & 0.079 &  \textbf{0.064} \\
            \hline
            Num test trajectories & 20 & 20 & 20 & 20 & 20 & 20 & 20 & 20  & 20 & 20   \\
            Length of test trajectories & 300 & 300  & 300 & 300 & 300 & 300 & 300 & 300 & 300 & 300  \\
            Average test RMS translation error (m) & 0.385 & \textbf{0.237} & 0.384  & \textbf{0.238} & 0.385 & \textbf{0.238} & 0.384 & \textbf{0.240}& 0.384 & \textbf{0.235} \\
            Average test RMS rotation error (rad) & 0.067& \textbf{0.059} &  0.067  &  \textbf{0.063} & 0.067 & \textbf{0.063} & \textbf{0.067} &  0.069 & 0.067 & \textbf{0.066} \\
            \toprule
        \end{tabular}
    }
    \label{table:detailD1}
    \end{center}
\end{table}

\begin{table}[h!]
    \vspace{-10mm}
    \renewcommand\thetable{III}
    \caption{Dataset $D_2$ - Comparison between LEO and Ours (final training runtime and error, test error statistics).}%\textit{BF} indicates brute force matching.}
    \begin{center}
    \vspace{-3mm}
    \resizebox{1\linewidth}{!}{
        \begin{tabular}{c|c|c||c|c||c|c||c|c||c|c}
            \toprule
            \textbf{Num trajectory in training set} & \multicolumn{2}{c||}{1} & \multicolumn{2}{c||}{5} & \multicolumn{2}{c||}{10} & \multicolumn{2}{c||}{20} & \multicolumn{2}{c}{30} \\
            \hline
            \textbf{Algorithm} & LEO & Ours & LEO & Ours & LEO & Ours & LEO & Ours & LEO & Ours \\
            \hline
            Num training iterations & 100 & 100 & 100 & 100 & 100 & 100 & 100 & 100 & 100 & 100  \\
            Parallelization enabled during training & yes & \textbf{no} & yes & \textbf{no} & $\text{yes}^*$ & \textbf{no} & $\text{yes}^*$ & \textbf{no} & $\text{yes}^*$ & no\\
            Time per training iteration (min) & 0.934 & \textbf{0.004} & 3.773 & \textbf{0.022} & 4.403 & \textbf{0.038} & 4.901 &  \textbf{0.075}& 6.96 & \textbf{0.113}\\
            \hline
            Average training RMS translation error (m) & 1.129 & \textbf{0.715} &   1.179 &  \textbf{0.730} & 1.090 &  \textbf{0.731}   &  1.057 & \textbf{0.718} & 1.104 & \textbf{0.761}\\
            Average training RMS rotation Avg error (rad) & 0.443  &  \textbf{0.056} &  0.467 & \textbf{0.051} &  0.421 &  \textbf{0.055} & 0.407 &  \textbf{0.056}  &0.429 &  \textbf{0.045} \\
            \hline
            Num test trajectories & 20 & 20 & 20 & 20 & 20 & 20 & 20 & 20  & 20 & 20   \\
            Length of test trajectories & 300 & 300  & 300 & 300 & 300 & 300 & 300 & 300 & 300 & 300  \\
            Average test RMS translation error (m) & 1.132 & \textbf{0.728} & 1.153  &  \textbf{0.733} & 1.583 & \textbf{0.735} & 1.043 & \textbf{0.734} & 1.093 & \textbf{0.771} \\
            Average test RMS rotation error (rad) & 0.447 &  \textbf{0.051} &  0.468 & \textbf{0.050}&  0.419 &  \textbf{0.055} & 0.402 &  \textbf{0.056} & 0.427&  \textbf{0.045}\\
            \toprule
        \end{tabular}
    }
    \label{table:detailD2}
    \end{center}
\end{table}
\vspace{-3mm}
For LEO, we faced compute memory problems on larger training sets (indicated by an asterix in tables \ref{table:detailD1} and \ref{table:detailD2}) which required us to decrease the number of threads and samples (needed to compute the loss in eq. \ref{leoloss}). The values of the hyperparameters used per training set size are specified in table \ref{table:detailD3}.
\begin{table}[h!]
    \vspace{-4mm}
    \renewcommand\thetable{IIII}
    \caption{Hyperparameters used to train LEO per training set size}%\textit{BF} indicates brute force matching.}
    \begin{center}
    \resizebox{0.6\linewidth}{!}{
        \begin{tabular}{c|c|c|c|c|c}
            \toprule
            Num trajectory in training set & \multicolumn{1}{c|}{1} & \multicolumn{1}{c|}{5} & \multicolumn{1}{c|}{10} & \multicolumn{1}{c|}{20} & \multicolumn{1}{c}{30} \\
            \hline
            Number of threads & 4 & 4 & 3 & 2 & 2 \\
            \hline
            Number of samples generated per training trajectory per iteration & 10 & 10 & 8 & 4 & 4 \\
           
            \toprule
        \end{tabular}
    }
    \label{table:detailD3}
    \end{center}
    \vspace{-3mm}
\end{table}

\end{document}